%% file: paperforreview.tex
\documentclass[10pt,twocolumn,letterpaper]{article}

\usepackage[pagenumbers]{cvpr} 

\usepackage{graphicx}
\usepackage{amsmath}
\usepackage{amssymb}
\usepackage{booktabs}
\usepackage{multicol}

\usepackage[pagebackref,breaklinks,colorlinks]{hyperref}
\usepackage{tablefootnote}
\usepackage[table]{xcolor}  

\usepackage[accsupp]{axessibility}  

\usepackage{algorithm}

\definecolor{deemph}{gray}{0.6}
\definecolor{ourscolor}{gray}{.9}
\newcommand{\ours}[1]{\cellcolor{ourscolor}{#1}}

\usepackage[capitalize]{cleveref}
\crefname{section}{Sec.}{Secs.}
\Crefname{section}{Section}{Sections}
\Crefname{table}{Table}{Tables}
\crefname{table}{Tab.}{Tabs.}

\begin{document}

\title{ Dense Distinct Query for End-to-End Object Detection}

\author{Shilong Zhang\textsuperscript{\rm 1,3 *},  \space\space
        Xinjiang Wang\textsuperscript{\rm 2 *},\space\space
        Jiaqi Wang\textsuperscript{\rm 1},\space\space
        Jiangmiao Pang\textsuperscript{\rm 1}, \\
        Chengqi Lyu\textsuperscript{\rm 1},\space\space
        Wenwei Zhang\textsuperscript{\rm 4,1},\space\space
        Ping Luo\textsuperscript{\rm 3,1},\space\space
        Kai Chen\textsuperscript{\rm 1}\\
        \small
        \textsuperscript{\rm 1}Shanghai AI Laboratory  \space \space
        \small\textsuperscript{\rm 2} SenseTime Research \\
        \small\textsuperscript{\rm 3} The University of Hong Kong \space \space
        \small\textsuperscript{\rm 4} S-Lab, Nanyang Technological University  \\
    }

\twocolumn[{
\maketitle  
\vspace{-14mm}
\centering

\begin{figure}[H]
\hspace{6mm}
\hsize=\textwidth
\includegraphics[width=0.95\textwidth]{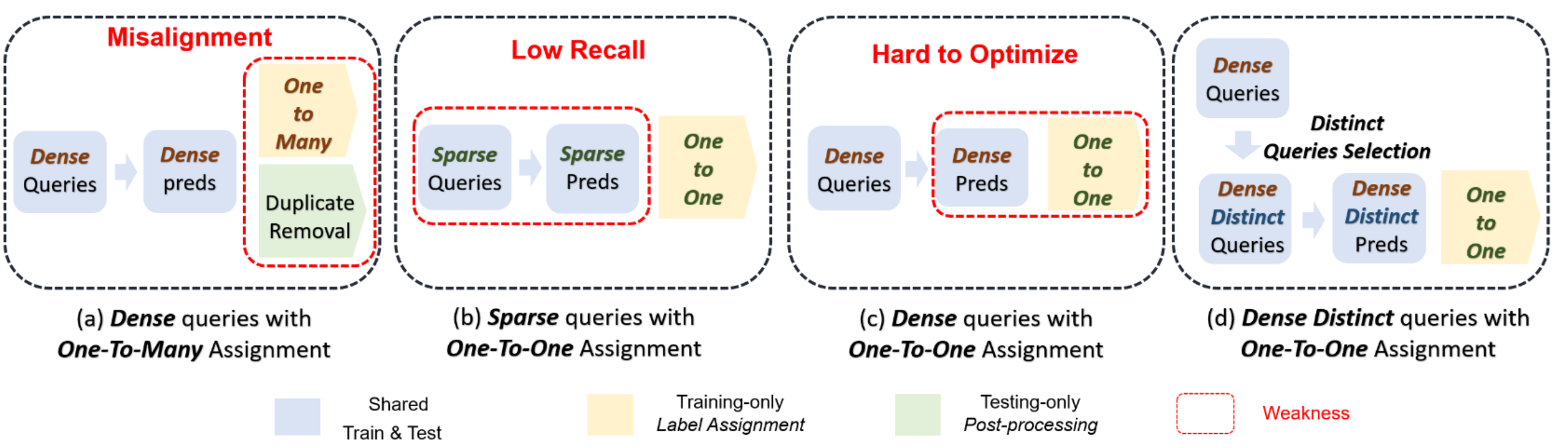} 
\caption{Pros and Cons of different queries and their corresponding learning paradigms. }
\label{fig:framework}
\end{figure}
}]

\footnotetext{* Equal contribution.}
\input{latex/sections/intro}

\input{latex/sections/related_works}
\input{latex/sections/method}
\input{latex/sections/results}

\section{Acknowledgement}
This project is supported by the National Key R\&D Program of China No.2022ZD0161600, No.2022ZD0161000 and the General Research Fund of HK No.17200622.

\section{Conclusions}

This paper reveals that both sparse and dense queries in end-to-end detection are problematic. We propose that the expected queries should be both dense and distinct. Such a paradigm significantly improves the performance
of various detectors including FCN, R-CNN, and DETRs. This proves the paradigm blends advantages from the traditional detectors and the recent end-to-end detectors. We hope it can inspire researchers to consider the complementarity between traditional methods and end-to-end detectors.

\clearpage
{\small
\bibliographystyle{ieee_fullname}
\bibliography{egbib}
}

\clearpage
\appendix
\input{latex/sections/sub}

\clearpage

\end{document}

%% file: latex/sections/intro.tex
\begin{abstract} One-to-one label assignment in object detection has successfully obviated the need for non-maximum suppression (NMS) as postprocessing and makes the pipeline end-to-end. However, it triggers a new dilemma as the widely used sparse queries cannot guarantee a high recall, while dense queries inevitably bring more similar queries and encounter optimization difficulties. As both sparse and dense queries are problematic, then what are the expected queries in end-to-end object detection? This paper shows that the solution should be Dense Distinct Queries (DDQ). Concretely, we first lay dense queries like traditional detectors and then select distinct ones for one-to-one assignments. DDQ blends the advantages of traditional and recent end-to-end detectors and significantly improves the performance of various detectors including FCN, R-CNN, and DETRs. Most impressively, DDQ-DETR achieves 52.1 AP on MS-COCO dataset within 12 epochs using a ResNet-50 backbone, outperforming all existing detectors in the same setting. DDQ also shares the benefit of end-to-end detectors in crowded scenes and achieves 93.8 AP on CrowdHuman. We hope DDQ can inspire researchers to consider the complementarity between traditional methods and end-to-end detectors.  The source code can be found at \url{https://github.com/jshilong/DDQ}.

\end{abstract}
\section{Introduction} \label{sec:intro} Object detection is one of the most fundamental tasks in computer vision, which aims at answering \emph{what objects are in an image and where they are.} To achieve the objective, the detector is expected to detect all objects and mark each object with only one bounding box.

Due to the complex spatial distribution and the vast shape variance of objects, detecting all objects is quite challenging. To solve the problem, traditional detectors ~\cite{ren2015faster, lin2017focal, tian2019fcos}  first lay predefined dense grid queries$^1$\footnote{1 Anchors ~\cite{ren2015faster, lin2017focal} or anchor points ~\cite{tian2019fcos}  in conventional detectors play the same role as sparse object queries in \cite{carion2020end}. Hence, we collectively refer to densely distributed anchor boxes and anchor points as dense queries.} to achieve a high recall. Convolutions with shared weights are then applied to quickly process dense queries in a sliding-window manner. At last, one ground truth bounding box is assigned to multiple similar candidate queries for optimization. However, the one-to-many assignment results in redundant predictions and thus requires extra duplicate-removal operations  (\emph{e.g.}, non-maximum suppression) during inference, which causes misaligned inference with training and hinders the pipeline from being end-to-end (as shown in Fig.~\ref{fig:framework}(a)).

This paradigm is broken by DETR~\cite{carion2020end}, which assigns only one positive query to each ground truth bounding box (one-to-one assignment) to achieve end-to-end. This scheme requires heavy computation to refine queries and adopts self-attention to model interactions between queries to facilitate the optimization of one-to-one assignment, which unfortunately limits the number of queries. For example, DETR only initializes hundreds of learnable object queries. Therefore, compared to the densely distributed queries in conventional detectors, the sparse queries fall short in recall rate, as shown in Fig.~\ref{fig:framework}(b).

Some recent works have also tried to integrate dense queries into one-to-one assignment~\cite{wang2021end, pmlr-v139-sun21b, yao2021efficient}. However, dense queries in end-to-end detectors face unique challenges. For example, our analysis shows that this paradigm would inevitably introduce many similar queries (potentially representing the same instance) and that it suffers difficult and inefficient optimization as similar queries are assigned opposite labels under one-to-one assignment.(Fig.~\ref{fig:framework}.(c)).

Now that both sparse queries (low recall) and dense queries (optimization difficulty) under one-to-one assignment are sub-optimal, \emph{what are the expected queries in
end-to-end object detection?}

In this study, we demonstrate that the solution should be \emph{dense distinct queries (DDQ)}, meaning that the queries for object detection should be both densely distributed to detect all objects and also distinct from each other to facilitate the optimization of one-to-one label assignment. Guided by such a principle, we consistently improve the performance of various detector architectures, including FCN, R-CNN, and DETRs. For one-stage detectors composed of fully convolutional networks (FCN), we first propose a pyramid shuffle operation to replace heavy self-attentions to model the interaction between dense queries. Then, a distinct queries selection scheme ensures that the one-to-one assignment is only imposed on the selected distinct queries, preventing contradictory labels from being assigned to similar queries.  Such an end-to-end one-stage detector is named \textbf{DDQ FCN} and achieves state-of-the-art performance in one-stage detectors. DDQ also naturally extends to DETR and R-CNN structures~\cite{zhu2020deformable, liu2021dab, dai2021dynamicdetr, sun2021sparse} by first laying dense queries as in~\cite{zhu2020deformable} and then selecting distinct queries for later refining stages, which are respectively dubbed \textbf{DDQ R-CNN} and \textbf{DDQ DETR}.

We have conducted experiments on two datasets---MS-COCO~\cite{lin2014microsoft} and CrowdHuman~\cite{shao2018crowdhuman}.  DDQ FCN and DDQ R-CNN obtain 41.5/44.6 AP, respectively, on the MS-COCO detection dataset ~\cite{lin2014microsoft} with the standard 1x schedule\cite{ren2015faster,lin2017focal}. Compared to recent DETRs, DDQ DETR achieved 52.1 AP in just 12 epochs with the DETR-style augmentation~\cite{carion2020end}.  The strong performance demonstrates that DDQ overcomes the optimization difficulty in end-to-end detectors and converges as fast as traditional detectors with higher performance. 

Object detection in crowded scenes such as CrowdHuman~\cite{shao2018crowdhuman} is another arena to testify to the effectiveness of DDQ. It is extremely cumbersome to tune the post-processing NMS in traditional detectors, as a low IoU threshold leads to missing overlapping objects, while a high threshold brings too many false positives. Recent end-to-end structures also struggle to distinguish between duplicated predictions and overlapping objects due to their difficult optimization. In this study, DDQ FCN/R-CNN/DETR achieve  92.7/93.5/93.8 AP and 98.2/98.6/98.7 recall on CrowdHuman~\cite{shao2018crowdhuman}, surpassing both traditional and end-to-end detectors by a large margin.

%% file: latex/sections/related_works.tex
\vspace{-5mm}

\section{Related Work}

\label{sec:related_work}
\noindent\textbf{Dense Queries with One-To-Many Assignment.}  One-stage detectors such as RetinaNet\cite{lin2017focal} and FCOS \cite{tian2019fcos} use densely distributed queries for regression and classification.The same manner is also applied to the region proposal network (RPN) of multi-stage models \cite{ren2015faster, cai2018cascade}. 
And one-to-many assignments are a common practice for these traditional detectors. Despite the fast development of one-to-many assignments from static label assignments (such as IOU-based~\cite{lin2017focal, ren2015faster, cai2018cascade} and center-based ones~\cite{tian2019fcos, kong2020foveabox}) to prediction-aware dynamic label assignments \cite{zhang2019freeanchor, zhu2020autoassign, ge2021ota, feng2021tood, li2022dual, chen2022diffusiondet}, these strategies are also long criticized for they pair each ground truth with multiple queries and thus require additional postprocessing to remove duplicate predictions at inference, which prevents the pipeline from being end-to-end. \\
\noindent\textbf{Sparse Queries with One-To-One Assignment.} DETR \cite{carion2020end} designs a small set of learned positional embeddings that represent the position in an image to focus on. These queries are then optimized with one-to-one assignments, making an end-to-end pipeline. Sparse R-CNN~\cite{sun2021sparse} reformulates queries in the traditional R-CNN framework as a bounding box and its corresponding embedding. Anchor DETR~\cite{wang2021anchor} provides the correspondence between anchor points and query position. DAB-DETR ~\cite{liu2021dab} explicitly learns a set of 4-D anchor boxes as queries. Though the formulation of queries varies, they share the same core idea of sparse queries and one-to-one assignments. Therefore, a low recall rate is an expected issue for these detectors. \\
\noindent\textbf{Dense Queries with One-To-One Assignment.} Both DeFCN \cite{wang2021end} and OneNet \cite{sun2021makes} try to integrate one-to-one assignment with dense queries. 
Despite their competitive performance compared to FCOS \cite{tian2019fcos}, there is still a clear performance gap with recent detectors with dynamic one-to-many assignment strategies~\cite{kim2020probabilistic, zhu2020autoassign, ge2021ota, feng2021tood, li2022dual}. It is the optimization difficulty of similar queries under one-to-one assignments that accounts for the performance gap. Efficient DETR~\cite{yao2021efficient}, and Two-Stage Deformable DETR \cite{zhu2020deformable} can also be regarded as a multi-stage version of this paradigm. Although DINO~\cite{zhang2022dino}, Group DETR~\cite{chen2022group}, and H-DeformableDETR~\cite{jia2022detrs} have introduced more positive samples to speed up convergence, the hindrance effect between similar queries and one-to-one assignments still remains unrevealed.

%% file: latex/sections/method.tex
\section{Analysis of Sparse and Dense Queries}

\begin{figure}[!h]
\centering
\includegraphics[width=0.75\linewidth]{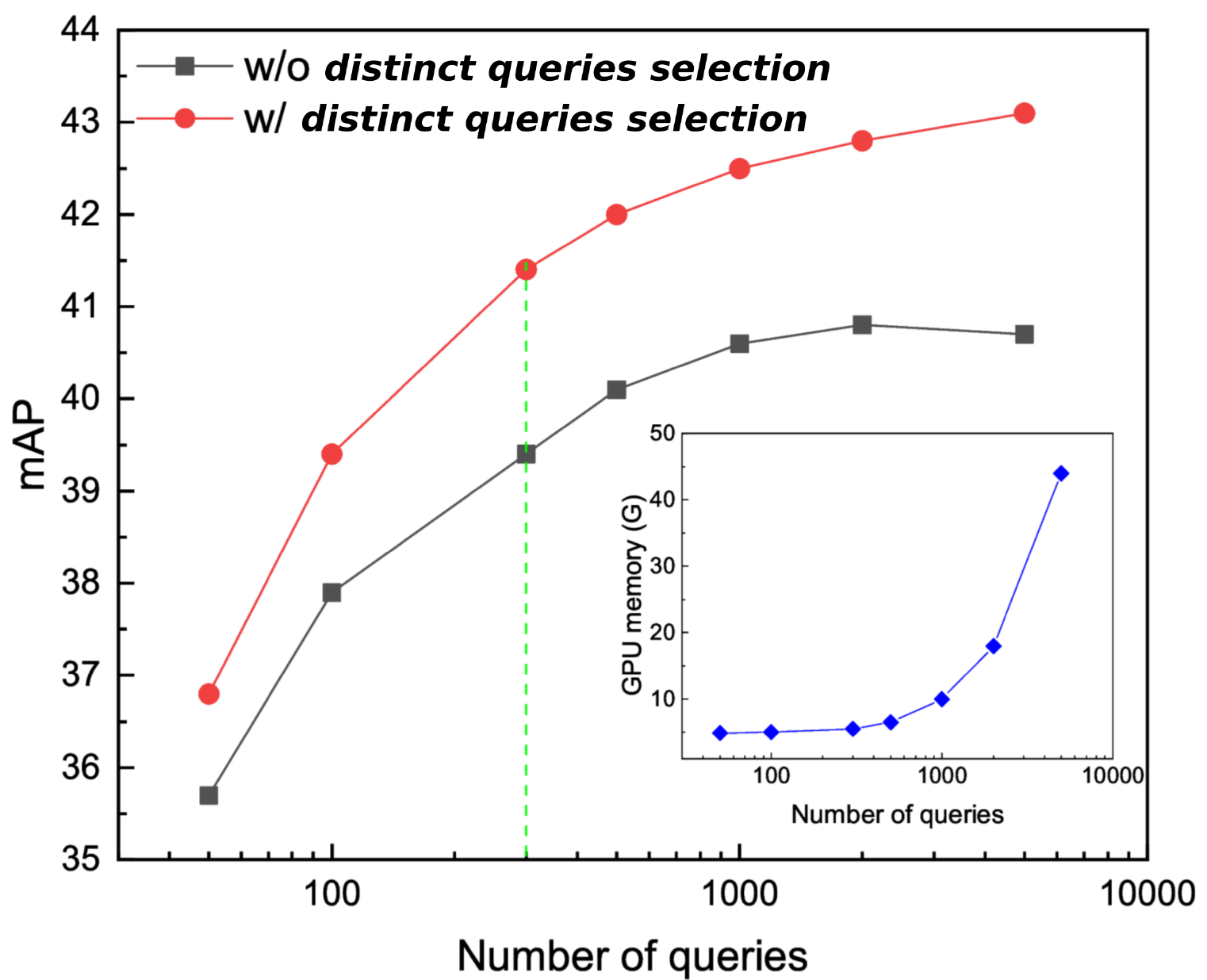}

\caption{The performance comparison of Sparse R-CNN with and without Distinct Queries Selection (a class-agnostic NMS with a threshold 0.7 before each refining head). All models are trained using the standard 1x setting. The green dotted line represents the default number(300) of queries adopted in Sparse R-CNN. The subplot denotes the memory consumption per GPU as the number of queries increases in Sparse R-CNN. 
}
\vspace{-5mm}
\label{fig:nms}
\end{figure}

Current end-to-end detectors use either dense or sparse queries, both of which are however problematic during training. 
Specifically, sparse queries suffer a low recall rate, and dense queries have issues in optimization. To illustrate this, we increase the number of queries from 10 to 7000 in Sparse R-CNN, and the performance is shown as the black line in Fig.~\ref{fig:nms}. The performance first keeps rising as the number of queries increases to around 2000, implying that the sparse queries ($\sim$ 300) in Sparse R-CNN are far from enough due to the low recall rate.  On the other hand, the performance finally plateaus and even decreases as queries number further increases. This phenomenon can be explained by the difficulty in distinguishing similar queries in end-to-end detectors with one-to-one assignment, especially when queries become denser. 

\label{sec.distincness}

To understand how similar queries would hinder optimization, we provide a simplified example where we assume there exist two identical queries. In this case, the one-to-one assignment assigns a foreground label to one of them but a background label to another. Without loss of generality, we adopt binary cross-entropy loss for classification. Therefore, the loss from these two queries becomes $L_1=-\log(p_1) - \log(1 - p_2)$, where $p_1$ and $p_2$ are the probability scores of the positive and negative query, respectively, and satisfy $p_1 = p_2 = p$ as they are identical queries. In contrast, the loss value when only one of the duplicated queries exists is $L_0=-\log(p)$. The ratio $\alpha$ of the gradient between the duplicate and non-duplicate query is.
\vspace{-2mm}
\begin{equation}
\alpha = \frac{\partial L_1}{ \partial p} / \frac{\partial L_0}{ \partial p}  = 1 - \frac{p}{1-p} \
\vspace{-1mm}
\end{equation}
It is obvious that the gradient is scaled down (\emph{i.e.}, $\alpha < 1$) at $0<p<0.5$ and may even cause negative training (\emph{i.e.}, $\alpha < 0$) at $p > 0.5$.

As shown in the toy example, duplicated queries reduce gradients and even cause negative training, which dramatically suppresses convergence. To avoid this issue, we impose a distinct queries selection operation before the one-to-one assignment process.  The distinct queries selection strategy is realized by a simple class-agnostic NMS. The filtered distinct queries are thus easier to optimize, and such an operation improves the performance by a clear margin, as seen from the red curve in Fig.~\ref{fig:nms}. More surprisingly, the performance margin consistently increases along with more queries. A similar trend is also observed for Deformable DETR, which can be found in the supplementary material. 

In other words, once we make sure the selected queries are distinct, the performance of Sparse R-CNN can be improved consistently with the increasing number of distinct queries. However, using a large number of distinct queries causes a significant memory footprint. For example, Sparse R-CNN requires around 45G memory per GPU with 7000 queries. To leverage the advantage of dense distinct queries(DDQ) with a reasonable computation cost, we give practical designs for all popular detector architectures (FCN, R-CNN, and DETRs).

\section{Method}

\begin{figure*}[t]
    \centering
    \includegraphics[width=\linewidth]{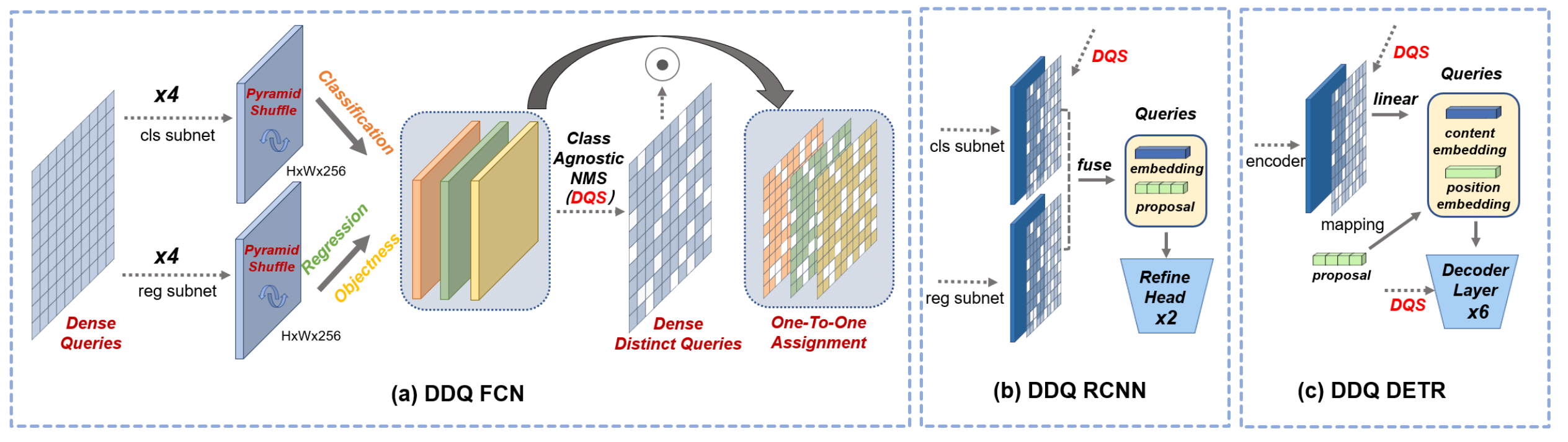}
      \caption{
    \textbf{The pipeline of DDQ}. (a) shows the application of DDQ to an FCOS-like structure, which is a fully-convolutional network (FCN). It is thus dubbed \textbf{DDQ FCN}. The pyramid shuffle is applied to the last two and last convolution layers in the classification and regression branches, respectively. The class-agnostic NMS act as the distinct queries selection operation. At last, only distinct queries will be assigned labels before calculating loss. (b) shows the design of DDQ for R-CNN structures (DDQ R-CNN). The last feature maps of the classification and regression branches of DDQ FCN are concatenated and filtered as distinct queries. The distinct queries are then sent to the refining heads with their corresponding bounding boxes. (c) shows the design of DDQ for DETRs  (DDQ DETR). After distinct queries selection, the remaining feature embedding in the encoder is projected with a linear to the content part of distinct queries. Their corresponding bounding boxes will be mapped to the position embedding part. Both parts will be sent to 6 refine stages. In such a long refine architecture, DQS will be applied before each refine stage to ensure distinctness.}
    
    \label{fig:pipeline}
    \vspace{-3mm}
\end{figure*}

Dense distinct queries (DDQ) is our principle for designing an object detector and can be integrated into different architectures. We first briefly describe the design of DDQ followed by detailed descriptions for the three architectures: FCN, R-CNN, and DETRs. The overall pipeline is sketched in Fig.~\ref{fig:pipeline}. 

\subsection{Paradigm of DDQ} 
\label{sec:method}
\noindent\textbf{Dense Queries}. As shown in Fig.~\ref{fig:nms}, the memory cost soars for dense queries. The main reason for this is the heavy calculation for each query. Instead of adopting learnable positional embedding in DETR, DDQ directly takes the feature point on each feature map as densely distributed initial queries. The number of queries in the feature pyramid can easily surpass 10000 given an input resolution of 800x800. 
To discriminate dense queries with reasonable computation cost, a light-weighted convolutional/linear network serves as the first stage and processes all queries in a sliding window manner.\\

\noindent\textbf{Distinct Queries} Now that the importance of query distinctness for optimization has been revealed in Sec.~\ref{sec.distincness}, we would discuss in this section why a class-agnostic non-maximum suppression (NMS) can be used to select distinct queries and how it differs from the traditional NMS as post-processing in traditional detectors. Since each query represents a potential instance in an image, and an instance can be uniquely represented by its location in an image~\cite{wang2020solo}, it comes naturally to detect similar queries using the class-agnostic overlapping ratio between the bounding boxes predicted by queries. More specifically, we apply a class-agnostic NMS to select distinct queries for the following one-to-one assignment. The loss is thus only computed on the selected distinct queries. \emph{It should be noted that such an operation is adopted in both training and inference, instead of only in inference as an extra post-processing in traditional detectors. Therefore, such a pipeline still abides by the definition of end-to-end detectors.} Compared to the training-unaware NMS in traditional detectors, it is designed to relieve the burden of one-to-one assignment during training, and can thus be set with an aggressive IoU threshold (0.7 in DDQ FCN and DDQ R-CNN, 0.8 in DDQ DETR), which is robust even on CrowdHuman dataset ~\cite{shao2018crowdhuman}. Such crowd scenes can not be properly handled by NMS as post-processing in traditional detectors. We validate this in Table.~\ref{tab:crowdhuman}.\\

\noindent\textbf{Loss Components} \label{sec:mainloss} \\
\emph{(1). Main Loss for Dense Distinct Queries}. We simply apply the bipartite matching algorithm in DETR~\cite{carion2020end} with the same cost weight in the one-to-one assignment. No extra prior (such as center priors in ~\cite{wang2021end}) is adopted for a fair comparison with DETRs. After discriminating positive and negative samples, DDQ FCN adopts GIoU loss~\cite{rezatofighi2019generalized} and QFocal loss~\cite{li2021generalized} with weights 2 and 1. For DDQ R-CNN and DDQ DETR, we just follow the implementation of Sparse R-CNN~\cite{sun2021sparse} and DINO~\cite{zhang2022dino}.

\noindent\emph{(2). Auxiliary Loss for Dense Queries}. Despite the more efficient optimization in DDQ due to the removal of similar queries, it also results in numerous "leaf" queries through which no gradients are back-propagated. Therefore, we design an auxiliary head and an auxiliary loss to further harness the potential of the filtered queries following the design in DeFCN~\cite{wang2021end}. The auxiliary head is mostly identical to the main head, except that it adopts a soft one-to-many assignment for dense queries to allow for dense gradients and more positive samples to speed up training. More details can be found in our supplementary material.
\vspace{-1mm}
\subsection{DDQ FCN}   As shown in Fig.~\ref{fig:pipeline}. (a), the DDQ principle is first applied to FCOS as an example of the FCN structure for object detection. It is found that dense queries are already available on the dense feature pyramid. However, as the dense queries are processed level by level with convolutional layers. The missing interaction across different levels poses a challenge for the optimization of one-to-one assignments.

\begin{figure}[!h]
\centering
\includegraphics[width=0.75\linewidth]{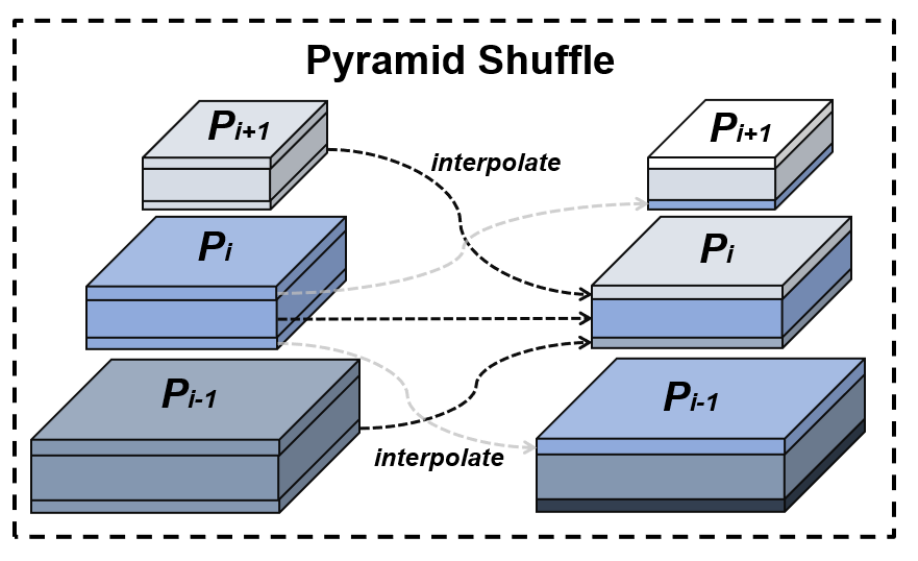}
\vspace{-3mm}
\caption{\textbf{An illustration of pyramid shuffle.} For queries in a specific scale level,  it can do the interaction with queries in an adjacent level by shuffling $S$ channels. Before concatenating to the feature of the target level, the feature from other levels should be interplate to the same size as the target level.}
\label{fig:shuffle}
\vspace{-3mm}
\end{figure}

Inspired by channel shuffle operation in ShuffleNet~\cite{zhang2018shufflenet}, we propose a pyramid shuffle to compensate for the interaction between queries in different levels where $S$ channels across adjacent levels are shuffled to form a new feature pyramid. Specifically, features at level $i$ exchange $S$ channels with those at level $i-1$ and level $i+1$ simultaneously. To account for the different spatial dimensions on the feature pyramid, a bilinear interpolation is adopted when exchanging features. We apply the pyramid shuffle operation on the last two and one convolution layer in the classification and regression branches, respectively. This approach stabilizes training and improves performance with negligible additional computation costs. In this work, we set $S$ to 64 which means each feature level exchanges information from 128 channels with other levels. (Comparison with other approaches to model the interaction among dense queries, ablation, and the analysis of pyramid shuffle can be found in our supplementary material.) 

As for the distinct queries selection module in DDQ FCN, we first select the top 1000 predictions according to the classification score from each feature level and then apply a class-agnostic non-maximum suppression with a threshold of 0.7 to ensure both distinctness and generality across different datasets.

\subsection{DDQ R-CNN} We combine DDQ FCN with two refine stages in Sparse R-CNN to construct the DDQ R-CNN.  As shown in Fig.~\ref{fig:pipeline}. (b), thanks to the fast processing of dense distinct queries in DDQ FCN, we select 300 most representative queries according to the classification score from the remaining distinct queries. Then we concatenate the feature in the distinct position of the last feature map of the classification branch and regression branch to construct the query embedding. The query embedding and the corresponding bounding box prediction will be passed to the refinement head of Sparse R-CNN. Different from Sparse R-CNN which requires 6 stages of iterative query refinement, DDQ R-CNN needs as few as 2 refinement stages. Actually, the long iteration stages in Sparse R-CNN mainly compensate for the drawbacks caused by the sparse and sometimes similar input queries. For one thing, sparse queries could not cover all instances at initialization and thus need long cascading stages to refine. For another, similar queries also require long refinements to distinguish from each other to output a one-hot prediction for each instance ~\cite{carion2020end}. In contrast, the dense distinct queries from DDQ R-CNN have addressed the above issues, and hence the number of iterative refinements can be significantly reduced. We also report the results when we change the number of queries and refinement heads of DDQ R-CNN in the supplementary material. 

\subsection{DDQ DETR} We construct DDQ DETR based on Deformable DETR*~\footnote{* indicates it is an improved version based on techniques in DINO~\cite{zhang2022dino}. Details can be found in our supplementary material}. As shown in Fig.~\ref{fig:pipeline}. (c), We follow Two-Stage Deformable DETR~\cite{zhu2020deformable} to process dense queries. Instead of initializing the content part with transformed coordinates, we fuse the feature map embedding of distinct positions as the content part, which makes the initial queries more distinct. A class-agnostic NMS with a threshold of 0.8 is set to select distinct queries before each refining stage. To compare with recent DETRs, we keep the original 6 refining stages and select $K$ distinct queries for the refining stages. We also select the top 1.5$K$ queries directly according to classification scores as dense queries for the auxiliary head in the decoder. The parallel forward of dense queries and distinct queries follows the H-DeformableDETR~\cite{jia2022detrs} and Group DETR\cite{chen2022group}. We set $K$ to 900, following DINO~\cite{zhang2022dino}.

%% file: latex/sections/results.tex
\section{Experiments}
\label{sec:results}
In this section, we first introduce two standard benchmarks MS COCO~\cite{lin2014microsoft} and CrowdHuman~\cite{shao2018crowdhuman}. Then we introduce the setting of training and inference on both datasets. We also present three examples to show how  end-to-end detectors with different architectures evolve to our DDQ step by step. At last, we compare DDQ with state-of-the-art conventional detectors and recent end-to-end detectors on MS COCO and CrowdHuman, which show that DDQ blends the advantages of two design paradigms. The latency of current popular models and DDQ is compared in our supplementary material.

\subsection{Datasets}
MS COCO 2017~\cite{lin2014microsoft} detection dataset is mainly used for comparison and ablation studies. It contains 118k training, 5k validation images, and 20k test images without annotations. There are on average 7 instances per image in this dataset. We report bounding box mean average precision (AP) as the performance metric, which is the mean average precision over multiple thresholds. If not specified, AP on the validation set is set as default. 

Besides, we also report the performance on the CrowdHuman dataset~\cite{shao2018crowdhuman}, which has 15k training images and 4.4k validation images with around 23 heavily occluded instances per image. For evaluation, we use AP, mMR, and Recall as the metrics. mMR means the average log miss rate over false positives per image ranging in $\left[10^{-2}, 10^0\right]$ following the official report \cite{shao2018crowdhuman}. A lower value of mMR means a better quality of high-scoring bounding boxes. All evaluation results are reported on the CrowdHuman validation subset. 

\subsection{Setting}
\noindent\textbf{COCO} ResNet-50~\cite{he2016deep} is the default backbone in this study if not specified. Most models adopt the 1x(12 epochs) training protocol in MMDetection~\cite{chen2019mmdetection}. AdamW~\cite{loshchilov2018decoupled} optimizer is used. For DDQ FCN, we set the initial learning rate to \textbf{$5\times 10^{-5}$} and weight decay to 0.1. For DDQ R-CNN, we used a learning rate of \textbf{$10^{-4}$} and weight decay of 0.05. The learning rate for both two CNN-based detectors decayed with a ratio of 0.1 at epoch 9 and epoch 12. For DDQ DETR, we utilized a learning rate of \textbf{$2\times 10^{-4}$} and a weight decay of 0.05, and the learning rate decayed with a ratio of 0.1 at epoch 12 only. To ensure a fair comparison with other studies, we classified our data augmentation into three types \label{link:aug}: \emph{Normal}, \emph{Multi-Scale}, and \emph{DETR}. \emph{Normal} augmentation rescaled images to a short side of 800 pixels, with only random flips applied. For \emph{Multi-Scale} augmentation, we used the classic multi-scale training range (480--800). Finally, the \emph{DETR} augmentation followed that of the study by Carion et al.~\cite{carion2020end}. 

\noindent\textbf{CrowdHuman} ResNet-50~\cite{he2016deep} is the default backbone. All conventional detectors and DDQ  adopt the 3x (36 epochs) schedule with multi-scale training~(480-800). All optimizer-related parameters are consistent with the setting on COCO. For end-to-end detectors with sparse queries, we follow the schedule(50 epochs) in Sparse R-CNN due to their slow convergence. The max detected instance number is changed to 500 for all conventional detectors following the \cite{wang2021end}. For a fair comparison, We increase the number of queries to 500 for DDQ FCN/R-CNN, Sparse R-CNN, and Deform DETR. For DDQ DETR, we just keep the same 900 queries as COCO.

\subsection{Evolving to DDQ}
In this section, we show how  detectors of different architectures evolve to DDQ. We can validate the importance of both density and distinctness from such a progressive development.

\noindent\textbf{From FCOS* to DDQ FCN} In Table.~\ref{tab:fcos-ddq},  We start from an FCOS equipped with the bipartite matching algorithm in DETR~\cite{carion2020end} and our main loss components mentioned in Sec.~\ref{sec:mainloss}, which is denoted as FCOS*. We adopt the \emph{normal} augmentation that is mentioned in \ref{link:aug} and train the model for 12 epochs. Due to the lack of cross-level interaction, its performance is quite unstable and fluctuates between 24.5 AP and 36.5 AP in a few successful experiments. We select the best result 36.5 as our baseline. After adding pyramid shuffle operations to interact with cross-level queries, the training becomes stable and gets 1.1 AP improvement with only 0.2 G flops and 0.2 ms latency increase. Adding a distinct queries selection operation boosts the performance from 37.6 AP to 40.6 AP with only 0.3 ms latency. Such a 3 AP improvement demonstrates that the distinctness of queries is vital for the one-to-one assignment. After adding an auxiliary loss for dense queries following DeFCN~\cite{wang2021end}, we get DDQ FCN with a state-of-the-art performance of 41.5 AP. DQS on the strong baseline (equipped with pyramid shuffle and
auxiliary loss) can be found in Table.~\ref{tab:distinct}. DQS still improves 2 AP (from 39.5 to 41.5).
\begin{table}[!h]
    \centering
    \vspace{-2mm}
    \caption{\textbf{From FCOS* to DDQ FCN.} PS stands for pyramid shuffle, and DQS means distinct queries selection operation. We also report the latency(L) and the flops(F).}
     \scalebox{0.90}{
    \begin{tabular}{l|c|c|c|c|c}
    \hline
        Method  & AP & AP$_{50}$ & AP$_{75}$ & L(ms) & F(G) \\
        \hline
        FCOS*  &36.5 & 54.4 & 40.3 & 21.9 & 200.5 \\
        \hline
        + \textbf{PS}  & 37.6  & 56.3 & 41.3& 22.1 & 200.7 \\
        \hline
        +\textbf{DQS}  & 40.6  & 60.3 & 44.5 & 22.4 & 200.7 \\
        \hline
        DDQ FCN  & 41.5 & 60.9 & 45.4 & 22.4 & 200.7  \\

\end{tabular}}
\vspace{-3mm}
\label{tab:fcos-ddq}
\end{table}

\noindent\textbf{From Sparse R-CNN to DDQ R-CNN} Table. \ref{tab:sparse-ddq} shows a progressive development from Sparse R-CNN to DDQ R-CNN. Sparse R-CNN with 300 queries achieves 39.4 AP within 12 epochs using the \emph{normal} augmentation that is mentioned in Sec.\ref{link:aug}. Increasing the number of queries to 7000 improves the performance to 40.6 AP, at the cost of a quite heavy detector. Applying a distinct queries selection at the beginning of each stage boosts the performance by 2.5 AP to 43.1 AP. At last, by replacing the first four refining stages with our DDQ FCN, which not only makes the structure more light-weighted but also allows even denser input queries, the performance further increases to 44.6 AP.

\begin{table}[!h]
    \centering
    \caption{\textbf{From Sparse R-CNN to DDQ R-CNN}. Q means the query, and DQS stands for the distinct queries selection }
     \scalebox{0.90}{
    \begin{tabular}{l|c|c|c|c|c}
    \hline
        Method & AP & AP$_{50}$ & AP$_{75}$  & L(ms) & F(G) \\
        \hline
        Sparse R-CNN  & 39.4 & 57.7 & 42.5  & 31.0 &160.2 \\
        \hline
        +7000Q  & 40.6 & 58.7  & 44.0 & 135.0 & 781.0\\
        \hline
        +\textbf{DQS}  & 43.1 & 62.6 &  47.1& 135.0 & 781.0 \\
        \hline
        DDQ R-CNN & 44.6 & 63.0 & 48.8 & 31.3 & 248.5  \\

    \end{tabular}}
\vspace{-3mm}
    \label{tab:sparse-ddq}
\end{table}

\noindent\textbf{From Deformable DETR* to DDQ DETR} Table \ref{tab:ddq-detr} illustrates the progressive development from Deformable DETR* to DDQ DETR. Deformable DETR* achieves 45.4 AP with 900 queries within 12 epochs using the DETR augmentation mentioned in Sec.\ref{link:aug}. By employing a linear layer to process the dense queries on the feature pyramid and constructing content parts with feature embeddings, the performance increases to 48.5 AP. However, initializing the content part as Two-Stage Deformable DETR(TS D-DETR) with mapped coordinates only achieves 46.7 AP, which is due to the lack of distinctness in the coordinates compared to the feature embedding. Adding an auxiliary loss for the decoder improves performance to 50.0 AP. Furthermore, by adding DQS before each refining stage, the performance further increases to 50.7 AP. Finally, by adding the P2 feature and 100 CDN queries as in DINO~\cite{zhang2022dino}, we achieve an impressive 52.1 AP, surpassing all detectors in the same setting. We show  distinctness can be complementary to CDN in Table.~\ref{tab:distinct} and analyze the reason in our supplementary material.  

\begin{table}[!h]
    \centering
    \caption{\textbf{From Deformable DETR*(D-DETR)  to DDQ DETR}. TS D-DETR stands for the naive two-stage version. Dense means initializing the content part with feature embedding. DQS stands for distinct queries selection. AUX-Decoder means the auxiliary loss for dense queries in the decoder. The flops only has comparative meaning and does not contain custom cuda operators}
     \scalebox{0.90}{
    \begin{tabular}{l|c|c|c|c|c}
    \hline
        Method & AP & AP$_{50}$ & AP$_{75}$  & L(ms) & F(G) \\
        \hline
        D-DETR*  & 45.4 & 63.0 & 49.1  & 45 & 264 \\
        \hline
        TS D-DETR  & 46.7 & 64.5 & 50.8  & 46 & 269 \\
        \hline
        +\textbf{Dense}  & 48.5 & 66.2  & 52.7 & 47 & 270 \\
        \hline
        +AUX-Decoder  & 50.0 & 67.4 & 54.8 & 47 &  270 \\
        \hline
        +\textbf{DQS}  & 50.7 & 68.1 & 55.7 & 58 & 270 \\
        \hline
        DDQ DETR$_{5 scale}$ & 52.1 & 68.9 &  57.3 & 114 & 860 \\
        \hline

    \end{tabular}}
\vspace{-3mm}
    \label{tab:ddq-detr}
\end{table}

\subsection{Comparison with Other Detectors}

\noindent \textbf{Results on CrowdHuman}  We select some recent representative studies for comparison with DDQ on crowded scenes. It is seen that traditional detectors struggle between a low recall rate and a high false positive rate. Although DW~\cite{li2022dual} assignment is the recent state-of-the-art one-to-many assignment strategy and shows a clear increase in Recall compared to ATSS, it suffers from more serious false predictions and thus leads to a high mMR. The performance of such traditional detectors is limited by the post-processing NMS. In the supplementary material, we also show it can not be properly handled by adjusting the IoU threshold because it is training unaware.

End-to-end detectors can achieve a higher theoretical recall rate due to the removal of NMS as a post-process. However, a high recall is not guaranteed in Sparse R-CNN and Deformable DETR due to their sparse query design. Although DeFCN~\cite{wang2021end} achieves a better performance than other end-to-end methods by adopting dense queries, it is still difficult for DeFCN to distinguish between crowded objects and duplicated predictions(optimization difficulty) which affects the mMR.

\begin{table}[!h]

    \begin{center}
    \caption{Performance on CrowdHuman}
    \label{tab:crowdhuman}
     \scalebox{0.90}{
    \begin{tabular}{l|c|c|c|c}
    \hline
    Method &Epochs & AP$_{50}$& mMR & Recall \\
    \hline
    ATSS &36 & 89.6 & 44.4 & 95.9 \\
    DW &36 & 89.0 & 57.6 & 97.4 \\
    Cascade R-CNN &36  & 86.0 & 44.1 & 89.2 \\
    Sparse R-CNN & 50 & 89.2  & 48.3 & 95.9 \\
    Deform DETR & 50 & 89.1 & 50.0 &  95.3 \\
    DeFCN &36  & 91.0 & 46.5  & 97.9 \\
    \ours{\textbf{DDQ FCN} }&\ours{36 }&\ours{\textbf{92.7} }&\ours{\textbf{41.0} }&\ours{\textbf{98.2}} \\
    \ours{\textbf{DDQ R-CNN} }&\ours{36  }&\ours{\textbf{93.5} }&\ours{\textbf{40.4}}&\ours{\textbf{98.6}} \\
    \ours{\textbf{DDQ DETR} }&\ours{36  }&\ours{\textbf{93.8} }&\ours{\textbf{39.7}}&\ours{\textbf{98.7}} \\
    \end{tabular}}
    \end{center}
\vspace{-4mm}
\end{table}

In contrast, DDQ surpasses these detectors on all metrics by a clear margin. For one thing, DDQ leads in Recall due to the dense queries that could cover most objects. For the other, DDQ also achieves the lowest mMR, as a merit of the distinctness among queries so that the detector can better differentiate false predictions.

\begin{table*}[!h]
    \vspace{-5mm}
    \centering
        \caption{Results on COCO Dataset. For DW, the * means we have retrained it with the same augmentation(480-800) as other methods using official implementation.} 
                \vspace{-1mm}
                
            \scalebox{0.90}{
    \begin{tabular}{l|c|c|c|c|c|c|c|c|c}
    \hline
        Method & Backbone  &  Val$/$Test   & Epochs   & AP & AP$_{50}$ & AP$_{75}$  & AP$_s$ & AP$_m$ & AP$_l$ \\
        \hline
        \hline
        \emph{\textbf{Aug:DETR}} \\

        Cascade R-CNN~\cite{cai2018cascade}  & ResNet-50 & val  & 36 & 44.3 & 62.4 & 48 & 26.6 & 47.7 & 57.7\\
        DAB DETR\cite{liu2021dab} & ResNet-50 & val & 50 &  42.6 & 63.2 & 45.6 & 21.8 & 46.2 & 61.1 \\
        DN-DETR\cite{Li_2022_CVPR} & ResNet-50 & val & 50 &  44.1 & 64.4 & 46.7 & 22.9 & 48.0 & 63.4 \\
        Deformable DETR~\cite{zhu2020deformable}& ResNet-50 & val  & 50  & 46.2 & 65.2 & 50.0 & 28.8 & 49.2 & 61.7 \\
        Efficient DETR ~\cite{yao2021efficient} & ResNet-50 & val & 36 & 44.2  & 62.2 & 48.0 & 28.4 & 47.5 & 56.6 \\
        Sparse R-CNN~\cite{sun2021sparse}& ResNet-50 & val & 36 & 45.0 & 63.4 &48.2& 26.9& 47.2 &59.5 \\
        DINO$_{4 scales}$~\cite{zhang2022dino}& ResNet-50 & val & 36 & 50.9 & 69.0 & 55.3 & 34.6 & 54.1 & 64.6 \\
        DINO$_{5 scales}$~\cite{zhang2022dino}& ResNet-50 & val & 36 & 51.2 & 69.0 & 55.8 & 35.0 & 54.3 & 65.3 \\

        \ours{\textbf{DDQ FCN}}& \ours{ResNet-50 }& \ours{val }& \ours{36 }& \ours{\textbf{44.8} }& \ours{64.1 }& \ours{49.4 }& \ours{29.9 }& \ours{47.8 }& \ours{56.0} \\
       \ours{\textbf{DDQ R-CNN}}& \ours{ResNet-50 }& \ours{val }& \ours{36 }& \ours{\textbf{48.1} }& \ours{66.6 }& \ours{53.0 }& \ours{32.3 }& \ours{51.2 }& \ours{60.7} \\
       \ours{\textbf{DDQ R-CNN}$_{with\_encoder}$}& \ours{ResNet-50 }& \ours{val }& \ours{36 }& \ours{\textbf{51.0} }& \ours{69.0 }& \ours{56.0 }& \ours{34.0 }& \ours{54.4 }& \ours{64.6} \\
                \ours{\textbf{DDQ DETR}$_{4 scales}$}& \ours{ResNet-50 }& \ours{val }& \ours{24}& \ours{\textbf{52.0} }& \ours{69.5}& \ours{57.2}& \ours{35.2}& \ours{54.9}& \ours{65.9} \\
                     \ours{\textbf{DDQ DETR}$_{5 scales}$}& \ours{ResNet-50 }& \ours{val }& \ours{24 }& \ours{\textbf{52.8} }& \ours{69.9}& \ours{58.1}& \ours{37.4}& \ours{55.7}& \ours{66.0} \\
                     \hline

        Sparse R-CNN & ResNeXt-64x4d-101 & test-dev  & 36  & 46.9 & 66.3   &  51.2 & 28.6  & 49.2 & 58.7 \\
        Deformable DETR & ResNeXt-64x4d-101 & test-dev   & 50  & 49 & 68.5  & 53.2 & 29.7 & 51.7 & 62.8  \\
      \ours{\textbf{DDQ FCN}} & \ours{ResNeXt-64x4d-101} & \ours{test-dev }& \ours{36  }& \ours{\textbf{47.7} }& \ours{67.0  }& \ours{52.6 }& \ours{30.4 }& \ours{49.9 }& \ours{58.3} \\
    \ours{\textbf{DDQ R-CNN}} & \ours{ResNeXt-64x4d-101 }& \ours{test-dev }& \ours{36  }& \ours{\textbf{49.9} }& \ours{68.8  }& \ours{54.8 }& \ours{31.8 }& \ours{52.2 }& \ours{61.7} \\
    \hline
        Sparse R-CNN~\cite{sun2021sparse}& Swin-B & val & 36 & 50.8 & 70.4 & 55.6 & 33.9 & 53.7 & 65.9 \\
       \ours{\textbf{DDQ R-CNN}}& \ours{Swin-B }& \ours{val }& \ours{36 }& \ours{\textbf{52.8} }& \ours{72.2}& \ours{57.9 }& \ours{37.6 }& \ours{56.2 }& \ours{66.9} \\
    \hline
                     
        \hline
        H-DeformableDETR$_{4 scales}$& Swin-L & val & 36 & 57.6&  76.5&  63.2&  41.4&  61.7& 73.9 \\
                DINO$_{4 scales}$& Swin-L & val & 36 & 58.0 & 76.1 & 64.0 & 40.1 & 62.2 & 74.3  \\
                \ours{\textbf{DDQ DETR}$_{4 scales}$}& \ours{Swin-L }& \ours{val }& \ours{30}& \ours{\textbf{58.7}}& \ours{76.8}& \ours{64.5}& \ours{41.6}& \ours{62.9}& \ours{74.3} \\
            \ours{\textbf{DDQ DETR}$_{4 scales}$}& \ours{Swin-L }& \ours{test-dev }& \ours{30 }& \ours{\textbf{58.8} }& \ours{77.0 }& \ours{64.6 }& \ours{39.4}& \ours{62.1}& \ours{74.0} \\

        \hline
        \hline
        \emph{\textbf{Aug:Multi-Scale }} \\

        ATSS~\cite{zhang2020bridging} & ResNet-101 & test-dev & 24 & 43.6 & 62.1  &  47.4  &  26.1 &  47.0 & 53.6 \\
        PAA~\cite{kim2020probabilistic}  & ResNet-101 & test-dev & 24 & 44.8 & 63.3 & 48.7   & 26.5  & 48.8 & 56.3 \\

        OTA~\cite{ge2021ota} & ResNet-101 & test-dev & 24 & 45.3 & 63.5 & 49.3   & 26.9  & 48.8 & 56.1 \\
        DW* \cite{li2022dual} & ResNet-101 & test-dev & 24 & 45.8 & 64.6 & 49.6  & 27.3  & 48.9 & 57.0 \\
      \ours{\textbf{DDQ FCN}} & \ours{ResNet-101} & \ours{test-dev} & \ours{24} & \ours{\textbf{45.9}} & \ours{65.1} & \ours{50.7} & \ours{28.3} &\ours{48.6} & \ours{55.6} \\

    \end{tabular}}

    \label{tab:comparison_coco}
\end{table*}

\noindent \textbf{Results on COCO} We adopt heavier backbones and longer schedules to fairly compare with other detectors on COCO. As shown in Table.~\ref{tab:comparison_coco}, we get all the results from the original study except those marked with *. We divide the results into two parts according to the augmentation. The first part adopts the augmentations in DETR~\cite{carion2020end} and reports the results on COCO validation dataset. DDQ remains its advantage among end-to-end object detectors using different backbone structures. It is worth emphasizing that DDQ FCN without any refinement architecture can already surpass most end-to-end detectors. DDQ R-CNN surpasses these methods by a large margin with only two refining heads and without encoder architecture. The performance of DDQ R-CNN (R-50) can be further improved by adopting an encoder structure as in SEPC~\cite{Wang_2020_CVPR} or DyHead~\cite{dai2021dynamichead}. For example, It achieves an impressive 51.0 AP by adopting 6 blocks in DyHead as encoder structure, which is denoted as DDQ R-CNN$_{with\_encoder}$(details about this model can be found in supplementary material). DDQ DETR outperforms recent DETR with a clear margin using R-50 as its backbone. When adopting a Swin-L backbone, it also surpasses the SOTA method DINO~\cite{zhang2022dino} by 0.7 AP.

The second part adopts a multi-scale training (480-800) strategy for 24 epochs and reports the results on the COCO test-dev using ResNet-101, which is widely used by conventional detectors.

\section{Ablation study}
\subsection{The Recall Improvement of Dense Queries}
\begin{table}[!h]
    \centering
    \vspace{-0.15cm}
    \caption{Recall improvement of dense distinct queries. Q means queries and DQS stands for the distinct queries selection. L stands for the latency of the model.}
     \scalebox{0.90}{
    \begin{tabular}{l|c|c|c|c}
    \hline
        Method   & AR$_{100}$    & AR$_{200}$  & AR$_{300}$ & L(ms) \\
        \hline
        Sparse R-CNN  & 78.4  & 83.4  & 85.5 &  31.0 \\
        \hline
        7000 Q \& DQS & 88.6 & 92.3  & 93.6 & 135.0  \\

        \hline
        \ours{DDQ R-CNN} &\ours{88.5} & \ours{91.8} & \ours{93.2} & \ours{31.3} \\

    \end{tabular}}

    \label{tab:recall}
\end{table}
We analyze the recall of IoU threshold 0.5. As shown in Table.~\ref{tab:recall}, we report the recall of the 5th stage input queries of Sparse R-CNN to make a fair comparison with the input queries of the refinement head in DDQ R-CNN. It can be seen that the Sparse R-CNN with 300 queries has a significantly lower recall(10.2 AR$_{100}$) than that with 7000 queries. In DDQ R-CNN, the queries from the DDQ FCN achieve a comparable recall to 7000 queries but with much less latency.
\vspace{-2mm}

\subsection{ DQS with Different IoU Threshold }
In this section, we show the robustness of distinct queries selection(DQS) with different IoU thresholds.  As shown in  Table.~\ref{tab:distinct}, the performance of DDQ FCN/R-CNN is quite robust when the IoU threshold ranges from 0.6 to 0.8. The performance drops slightly when the threshold is lower than 0.6, which is due to the lower recall rate for overlapping objects. The performance also starts to degrade when the IoU threshold is larger than 0.8 due to its incapability to suppress similar queries that slow the optimization. DDQ DETR exhibits a similar trend, as observed in Table~\ref{tab:ddq-detr}. We can find even though CDN training in ~\cite{zhang2022dino} has been adopted in DDQ DETR, distinctness still improves the performance. By the way, we also report the performance of ATSS~\cite{zhang2020bridging} at different post-processing NMS IoU thresholds and show its sensitivity to this hyperparameter, in contrast to the robust behavior of DDQ in which a class-agnostic NMS is adopted in both training and inference to filter out distinct queries.

\begin{table}[!h]
 \vspace{-2mm}
    \begin{center}
         \caption{Performance of DDQ on COCO when DQS adopts different IoU thresholds. Results of ATSS adopting different IoU thresholds in post-processing are also reported. None means we remove DQS or post-processing from the inference pipeline.* means the results are not stable and we report the average performance}
    \label{tab:distinct}
\scalebox{0.90}{
    \begin{tabular}{l|c|c|c|c|c|c}
    \hline
         COCO   & 0.5 & 0.6 & 0.7 & 0.8 & 0.9 & None\\
      \hline
        DDQ FCN  &  40.8 & 41.4 & \ours{\textbf {41.5}}  & 41.4 & 40.5 & 39.5*  \\
         DDQ R-CNN & 44.0 & 44.5 & \ours{\textbf {44.6}} & 44.4 &  43.8  & 42.7* \\
        DDQ DETR & 50.1 & 50.7 & 50.9 & \ours{\textbf{51.3}} & 51.0 &   50.7*  \\
          
          ATSS  & 39.3 & \textbf{39.5} & 39.3 &38.7& 36.7 & 19.6\\

    \end{tabular}}
    \vspace{-5mm}
    \end{center}
\end{table}

\vspace{-2mm}

%% file: latex/sections/sub.tex
In the supplementary material, the analysis of dense distinct queries (DDQ) to Deformable DETR~\cite{zhu2020deformable} is first sketched in Sec.~\ref{sec:deformable_detr}. The details about auxiliary loss are added in Sec.~\ref{sec:loss}.   Sec.~\ref{sec:ana_py} gives more detailed ablation studies and analysis of pyramid shuffle.  Sec.~\ref{sec:stage} gives more detailed ablation studies about the number of queries and refining stages in DDQ R-CNN.   Sec.~\ref{sec:heavy} show the details about DDQ R-CNN with encoder. Sec.~\ref{sec:imporovedeform} elaborate and improved Deformable DETR and discuss the difference with DINO. Sec.~\ref{sec:latency} gives the latency benchmark.
Sec.~\ref{sec.query_form} reports the results of DDQ R-CNN with other ways to construct the query. Sec.~\ref{sec:nms} show the results of traditional detectors with different IOU thresholds on CrowdHuman. At last, Sec.~\ref{sec:social} illustrates our social impact.

\section{Analysis of DDQ in Deformable DETR}
\label{sec:deformable_detr}
For fast verification of Distinct Queries Selection(DQS) in such a heavy model, we adopt the standard 1x setting on COCO and keep other hyperparameters (such as learning rate and weight decay) the same with that in Deformable DETR~\cite{zhu2020deformable}.
\begin{figure}[!h]
    \centering
    \includegraphics[width=0.8\linewidth]{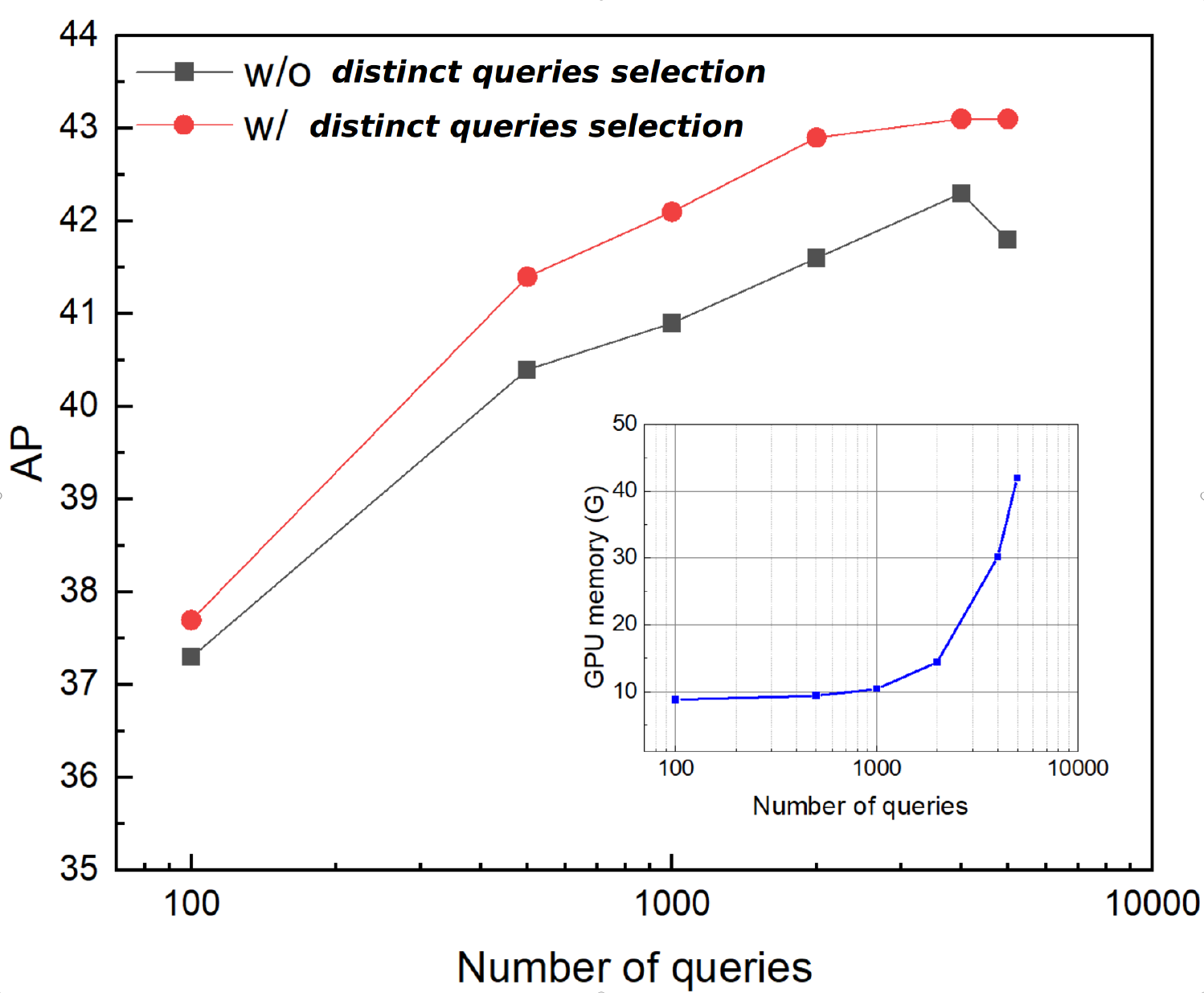}
    \caption{The performance comparison of Deformable DETR with and without distinct queries selection}
    \label{fig:nms_deform_detr}
\end{figure}

As shown in  Fig.\ref{fig:nms_deform_detr},  when we increase the number of queries in Deformable DETR, there is a similar trend as it in Sparse R-CNN~\cite{sun2021sparse}, as shown in the main manuscript. When the number of queries naively increases without distinct queries selection, the performance increases at the beginning but decreases as the number of queries reaches $\sim5000$. It is due to the more difficult training with more similar queries out of the dense queries. By imposing a distinct queries selection pre-processing to filter out similar queries and keeping only distinct queries before each stage of iterative refinement, the performance is improved with a clear margin, and the performance margin consistently increases along with more queries.

Therefore, we believe Dense Distinct Queries (DDQ) is a principle of designing an object detector with a fast convergence based on recent end-to-end detectors.

\section{Auxiliary Loss for Dense Queries}
\label{sec:loss}
We follow the  TOOD~\cite{feng2021tood}  to design our auxiliary loss. We select $K$ samples with the smallest cost of each ground truth as positive samples. $P$ means the index set of positive samples which correspond to the same ground truth. The classification score target of a sample $i$ in this set
is 
\begin{equation}
 \frac{ score_i * IoU_i ^{6}} {Max(score_j * IoU_j ^{6})_{j\in{P}}} * Max(IoU_j)_{j\in{P}}
\end{equation}

The GIoU loss of each sample is reweighted by the classification target. The classification loss and regression loss weight keep consistent with the main loss weight (1 and 2 respectively) for distinct queries. The performance is quite stable for DDQ FCN when $K$ ranges between 5 and 16. We adopt 8 and 4 for DDQ FCN and DDQ DETR respectively in this study. The auxiliary loss also works for refining heads in DDQ RCNN. Due to time issues, we will supplement relevant results in the future version.

\section{More Analysis and Ablation for Pyramid Shuffle}
\label{sec:ana_py}
\begin{figure*}[t]
    \centering
    \includegraphics[width=\linewidth]{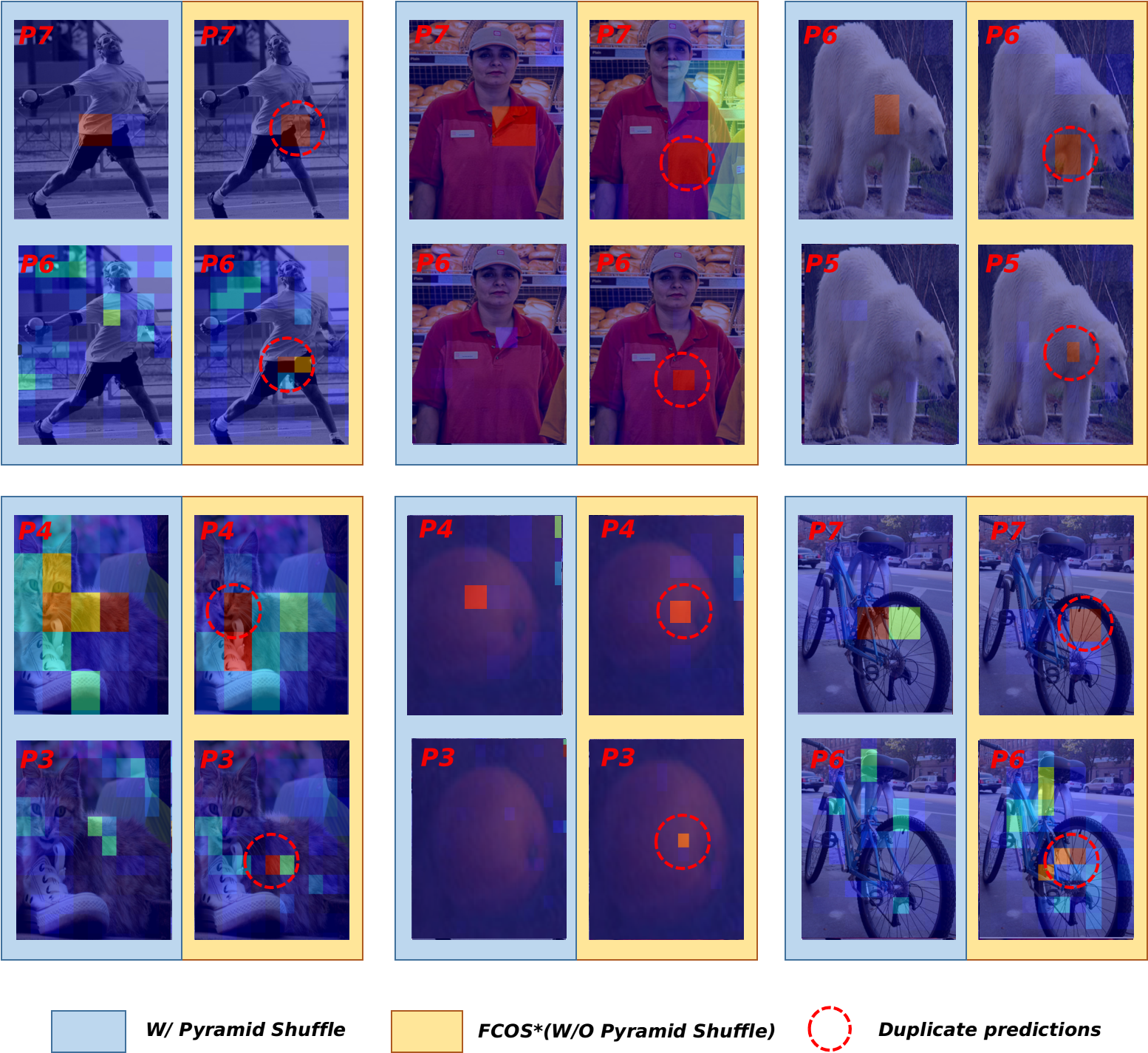}
      \caption{
    \textbf{Visualization of score map of adjacent levels}. We visualize classification scores with the rainbow color system.  The left side of each subfigure(with blue background) shows score maps with pyramid shuffles, and the corresponding right side (with yellow background) means without pyramid shuffles. The top-left corner marks the feature level. The red circle represents the duplication predictions in the adjacent level.}
    
    \label{fig:vis}
\end{figure*}

We provide an in-depth analysis of pyramid shuffle. Firstly, we compare it with 3D MAX Filter in DeFCN~\cite{wang2021end} in Table.~\ref{tab:3d}. Then we visualize the change of score maps in different levels after adding pyramid shuffle operations in Fig.~\ref{fig:vis}. At last, Table.~\ref{tab:pyshuffle} gives the results under different shuffle channels.

\begin{table*}[!h]
\vspace{-1mm}
    
    \centering
        \caption{Latency(ms) of different models with batch size 1}
        \vspace{-1mm}
        
    \begin{tabular}{c|c|c|c|c|c|c}
    \hline
      \ours{\textbf{ DDQ FCN}} & Cascade R-CNN  & Sparse R-CNN  & Deformable DETR  & \ours{\textbf{DDQ R-CNN} } & DINO & \ours{\textbf{DDQ DETR}}\\
                \hline
 \ours{~\textbf{44.8}} AP & 44.3 AP &  45.0 AP & 46.2  AP & \ours{ \textbf{48.1} AP } & {50.9} AP & \ours{ \textbf{52.0} AP }  \\
        \hline
          \ours{~\textbf{22.4} ms} &   28.5 ms   & 31.0 ms & 40.0 ms & \ours{\textbf{ 31.3}} ms  & 46 ms & \ours{\textbf{ 58}} ms

\label{tab:latency}
    \end{tabular}
    \vspace{-3mm}
\end{table*}

\begin{table}[!h]
    \begin{center}
\caption{\textbf{Comparison between 3D MAX Filter and Pyramid Shuffle.} * means the results is unstable}
    \label{tab:3d}
\scalebox{0.9}{ \begin{tabular}{l|c|c|c}
    \hline
        Flops & Parameters & Operations  & AP   \\
        
        \hline
         - & - & - & 41.0*   \\ 
        12.2 G & 0.59M & Conv\&GN\&Relu\&MaxPool3d & 41.2  \\ 
      \ours{0.2 G} & \ours{0.00M} & \ours{Shuffle x3}  & \ours{\textbf{41.5}} \\
        \ours{12.2 G} & \ours{0.59M} &  \ours{Conv\&GN\&Relu\& Shuffle x3}  & \ours{\textbf{42.0}} \\
         
    \end{tabular}}
    \end{center}

\vspace{-6mm}
\end{table}

\noindent\textbf{Comparision with 3D MAX Filter} Table.~\ref{tab:3d} shows the comparison between pyramid shuffle and 3D MAX Filtering in DeFCN. DeFCN believes there should be extra parameters and max pooling operation to facilitate the optimization under the one-to-one assignment. However, we argue that only the interaction of cross-level queries matters, and extra parameters or max operations are unnecessary. Our pyramid shuffle is more lightweight and with better performance. When we add an extra convolution to regression branches to fair compare with the 3D MAX Filter, we can surpass it by 0.8 AP.

\noindent\textbf{Visualization of Adjacent Level Score Map} Fig.~\ref{fig:vis} shows the scores map of adjacent levels. The left side of each subfigure(with blue background) shows score maps with pyramid shuffles, and the corresponding right side (with yellow background) means without pyramid shuffles. The top-left corner marks the feature level. The red circle represents the duplication predictions in the adjacent level. We can find pyramid shuffle effectively reduces the cross-level high score false positives.

\noindent\textbf{Results under Different Shuffle Channels} Table~\ref{tab:pyshuffle} gives the results under different shuffle channels; we can find that the number of channels can even be reduced to 16 when there is already cross-level distinct queries selection, making it more lightweight. When the number of shuffle channels is 128, which means no remaining channels for the current level, the performance will dramatically drop 1.5 AP because of missing information on the current level queries in the interaction.

\begin{table}[!h]
    
    \begin{center}
\caption{Performance of one-stage DDQ with different numbers of shuffle channels.}
    \label{tab:pyshuffle}

    \begin{tabular}{l|c|c|c|c}
    \hline
        Number   & AP  & AP$_{50}$ & AP$_{75}$ \\
        \hline
        0 &  41.0* & 59.9 & 45.1   \\
        8   & 41.2 & 60.3 & 45.4    \\
        16   & 41.3 & 60.6 & 45.4    \\
        32    & 41.4 & 60.6  & 45.5    \\
        \ours{64}    & \ours{\textbf{41.5}}  & \ours{60.9} & \ours{45.4}  \\
        96    &  41.6 & 61.1 & 45.7   \\
       128    & 39.9  & 59.9 & 43.6   \\
 
    \end{tabular}
    \end{center}

\vspace{-6mm}
\end{table}

\subsection{Number of Pyramid Shuffle Operations}
We report the results of DDQ FCN with the different number of pyramid shuffle operations in the classification and regression branches. When no pyramid shuffle is adopted in the DDQ FCN, its performance is unstable and fluctuates between 40.8 AP and 41.1 AP. We report an average performance of 41.0 AP. Even though there has been a cross-level distinct queries selection operation, compared to adopting only 2 and 1  operations to two branches respectively, there is still a 0.5 AP drop.

\begin{table}[!h]
    
    \begin{center}
    \caption{ Different number Pyramid shuffle operations in DDQ FCN. Cls means the classification branch and Reg means the regression branch. * indicate the performance is unstable}
    \label{tab:pyshuffle}
    
     \scalebox{0.90}{
    \begin{tabular}{l|c|c|c|c}
    \hline
        Cls & Reg   & AP  & AP$_{50}$ & AP$_{75}$ \\
        \hline
        0  & 0 &  41.0* & 59.9 & 45.1   \\
        1  & 0   & 41.3 & 60.1 & 45.6    \\
        0  & 1   & 41.3 & 60.6 & 45.4    \\
        0  & 2 & 41.3  & 60.0 & 45.6   \\
        2  & 0 & 41.2  & 60.1 &  45.5  \\
        1  & 1    & 41.3 & 60.6  & 45.8    \\
        \ours{2}  & \ours{1}    & \ours{\textbf{41.5}}  & \ours{60.9} & \ours{45.4}  \\
        2  & 2    &  41.4 & 60.6 & 45.5   \\
       4  & 4    & 41.5  & 61.1 & 45.6   \\
 
    \end{tabular}}
    \end{center}

\vspace{-6mm}
\end{table}

\vspace{-2mm}

\section{DDQ R-CNN with Encoder}
\label{sec:heavy}
The encoder can provide a more powerful feature representation for the decoder head which is explored in~\cite{zhu2020deformable,dai2021dynamicdetr, Wang_2020_CVPR}. We simply add 6 dynamic blocks in DyHead~\cite{dai2021dynamichead} as our encoder. 500 queries and 3 refinement stages are adopted. It is observed in the main manuscript that there is about 3 AP improvement for DDQ R-CNN.

\section{Details of Improved DeformableDETR and Comparison with DINO}
\label{sec:imporovedeform}
DINO~\cite{zhang2022dino} adopt some techniques that significantly improve the Deformable DETR. We remove the CDN and mix query selection from DINO to form our baseline. DDQ is a concurrent work of DINO. The contrastive denoising training (CDN) is not
intended to relieve the optimization difficulty of very similar
queries among dense queries. In their implementation, the
generated positive and negative samples in each pair are
always significantly distinct from each other. Mix query
selection increases the distinctness of queries by additionally initializing content embeddings, but the position embeddings are still created from top-k dense regression predictions which can be very similar and still hinder the
optimization. We have shown our components can be combined with DINO.

\section{Number of Queries and Stages in DDQ R-CNN }
\label{sec:stage}
\begin{table}[!h]
    \begin{center}
     \caption{Performance with different number of refine stages(S) and queries(Q).}
    \label{tab:queryandstage}
     \scalebox{0.90}{
    \begin{tabular}{l|c|c|c|c}
    \hline
          & 100 Q   & 200 Q & 300 Q & 400 Q \\
      \hline
        S=1  & 43.0  &  43.2  & 43.4 & 43.2   \\
      
        S=2  & 44.2  & 44.2  & \ours{\textbf{44.6} }    & 44.8    \\
   
        S=3  & 44.4  &  44.8 & 45.1  &  44.9       \\
 
        S=4  & 44.5   & 45.0  & 45.0 &    45.2   \\

    \end{tabular}}
    
    \vspace{-7mm}
    \end{center}
\end{table}
We analyze the combination of different numbers of stages and queries for DDQ R-CNN. Table.~\ref{tab:queryandstage} shows that the best number of stages is proportional to the number of queries. This is easy to understand. When the number of queries increases, the newly added queries are of low quality, and more stages are needed to refine these queries. It is worth emphasizing that we use 2 stages and 300 queries to trade off the performance and latency. When using 3 stages with the same number of queries, our method achieves an even higher performance of 45.1 AP on MS COCO.

\section{Latency Benchmark}
\label{sec:latency}
As for the details of latency calibration, We compare the speed (forwarding + post-processing) of  different methods  with batch size 1 in Table.~\ref{tab:latency}. All evaluations were performed on Tesla A100 GPU with Intel(R) Xeon(R) Gold 6348 CPU @ 2.60GHz. The Pytorch version is 1.9.0 with CudaToolkit 11.1 and Cudnn 8.0.5. An average of 200 iterations during model inference is adopted as the latency reported in this study.

\section{Impact of Query Construction in DDQ R-CNN }
\label{sec.query_form}

We try five ways to construct queries in DDQ R-CNN. As shown in Table. \ref{tab:constrctquery}, None means all query embeddings are set to a zero tensor, and the refinement stages only get meaningful query bounding boxes. This attempt reduces the performance to 43.2 AP. Simply constructing queries from the FPN results in 1.0 AP degradation. Reg means only using the last feature map of the regression branch, which drops the performance by 0.6 AP.  Constructing queries from the last feature map in the classification branch can be an alternative as it can get a comparable performance(only 0.3 AP drop).

\vspace{-3mm}
\begin{table}[!h]
    \centering
    \caption{Impact of Query Construction in DDQ RCNN}
     \scalebox{0.90}{
    \begin{tabular}{l|c|c|c}
    \hline
            &AP  & AP$_{50}$ & AP$_{75}$   \\
        \hline
        None &  43.2 & 61.0 & 47.9   \\
        FPN & 43.6 & 61.7 & 48.0      \\
        Cls & 44.3 & 62.2 & 48.6 \\
        Reg & 44.0 &  62.5 & 48.3\\
        \ours{Cls\&Reg}  & \ours{\textbf{44.6}} & \ours{63.0} & \ours{48.8}
    \end{tabular}}
    \label{tab:constrctquery}
    \vspace{-3mm}
\end{table}

\section{DQS with Different IoU Threshold in CrowdHuman }
\label{sec:nms}
In this section, we show the robustness of distinct queries selection(DQS) with different IoU thresholds in CrowdHuman. We can find there is a clear performance bottleneck for traditional detector ATSS even with carefully adjusting the threshold of NMS.

\begin{table}[!h]

    \begin{center}
         \caption{Performance of DDQ on COCO  when DQS adopts different IoU thresholds. Results of ATSS adopting different IoU thresholds in post-processing are also reported. None means we remove DQS or post-processing from the inference pipeline.}
    \label{tab:distinct}
\scalebox{0.90}{
    \begin{tabular}{l|c|c|c|c|c|c}
    \hline

          \hline
           CrowdHuman  & 0.5 & 0.6 & 0.7 & 0.8 & 0.9 & None \\
           \hline
        DDQ FCN  &  88.0 & 91.8  & \ours{\textbf {92.7}}  & 92.8 & 92.3 & 91.7  \\
         DDQ RCNN   & 91.8  & 92.9 & \ours{\textbf {93.5}} & 93.3 & 93.2  & 93.2 \\
          ATSS  & 88.3 &\textbf{89.6} & 88.4 & 85.3 & 78.9 & 42.7

    \end{tabular}}

    \end{center}
\end{table}

\section{Social Impact}
\label{sec:social}
The potential social impact of this work inherits from object detection. Because human behaviors often cause crowded scenes, and DDQ achieves excellent performance in such scenes, it may be applied to some applications that violate human privacy, such as surveillance.

\clearpage
%
%